\documentclass[letterpaper, 10 pt, conference]{ieeeconf}  

\IEEEoverridecommandlockouts                              
\usepackage{graphicx}
\usepackage{subfig}
\usepackage{amsmath} 
\usepackage{amssymb}  
\usepackage{multirow}
\usepackage{multicol}
\usepackage{balance}

\usepackage{tikz}
\usepackage{booktabs}
\usepackage{comment}

\newcommand{\DKL}{\mathbb{D}_{\mathrm{KL}}}
\newcommand{\para}[1]{\vspace{0.5em}\noindent\textbf{#1} }

\title{\LARGE \bf
Multi-Modal Mutual Information (MuMMI) Training for\\Robust Self-Supervised Deep Reinforcement Learning
}

\author{Kaiqi Chen$^*$, Yong Lee$^*$, and Harold Soh\\Dept. of Computer Science, National University of Singapore.\\{\small\texttt{\{kaiqi, liy, harold\}@comp.nus.edu.sg}}
\thanks{$^*$Equal Contribution.}
}

\begin{document}

\maketitle
\thispagestyle{empty}
\pagestyle{empty}

\begin{abstract}

This work focuses on learning useful and robust deep world models using multiple, possibly unreliable, sensors. We find that current methods do not sufficiently encourage a shared representation between modalities; this can cause poor performance on downstream tasks and over-reliance on specific sensors. As a solution, we contribute a new multi-modal deep latent state-space model, trained using a mutual information lower-bound. The key innovation is a specially-designed density ratio estimator that encourages consistency between the latent codes of each modality. We tasked our method to learn policies (in a self-supervised manner) on multi-modal Natural MuJoCo benchmarks and a challenging Table Wiping task. Experiments show our method significantly outperforms state-of-the-art deep reinforcement learning methods, particularly in the presence of missing observations.  

\end{abstract}
\section{Introduction}
\label{sec:intro}

We live in a rich complex world. To make sense of it,  humans (and other biological organisms) integrate information from a variety of senses. Our sensory apparatus (e.g., eyes, ears, skin) are often complementary, but also provide redundant information. This redundancy promotes robustness; biological agents display the incredible ability to cope under the temporary, or even permanent, loss of any given sense.

One might expect that artificial agents and robots can reap similar benefits from multiple sensory modalities. Indeed, many modern-day robots are equipped with a variety of sensors---e.g., cameras, microphones, tactile and proprioception sensors---that enable them to better perceive their environment. When combined with powerful representation learners (such as deep neural networks), these different sources of information can be used to learn world models for more robust decision-making and policy learning. 

Unfortunately, learning robust world models from multiple raw sensory inputs remains challenging. Rather than improving performance, our preliminary deep reinforcement learning (RL) experiments revealed that including additional modalities can cause performance to \emph{deteriorate}. The learned policies often failed to match the performance of a single-modality model, and were not robust to missing data. 

In this work, we address the issue above and answer the question: \emph{how can we learn complex world models from multiple, but possibly unreliable, sensors?} We develop a \emph{modular} multi-modal deep latent state-space model (MSSM) that can be used for various robot tasks, including model-based RL and planning. 
Compared to deep models that ``concatenate'' different modalities~\cite{zambelli2020multimodal,lee2019feifeili1}, structural modularity in our  probabilistic graphical model provides a principled technique for dealing with missing data (rather than masking) with fewer parameters.

\begin{figure}
    \centering
    \includegraphics[width=\columnwidth]{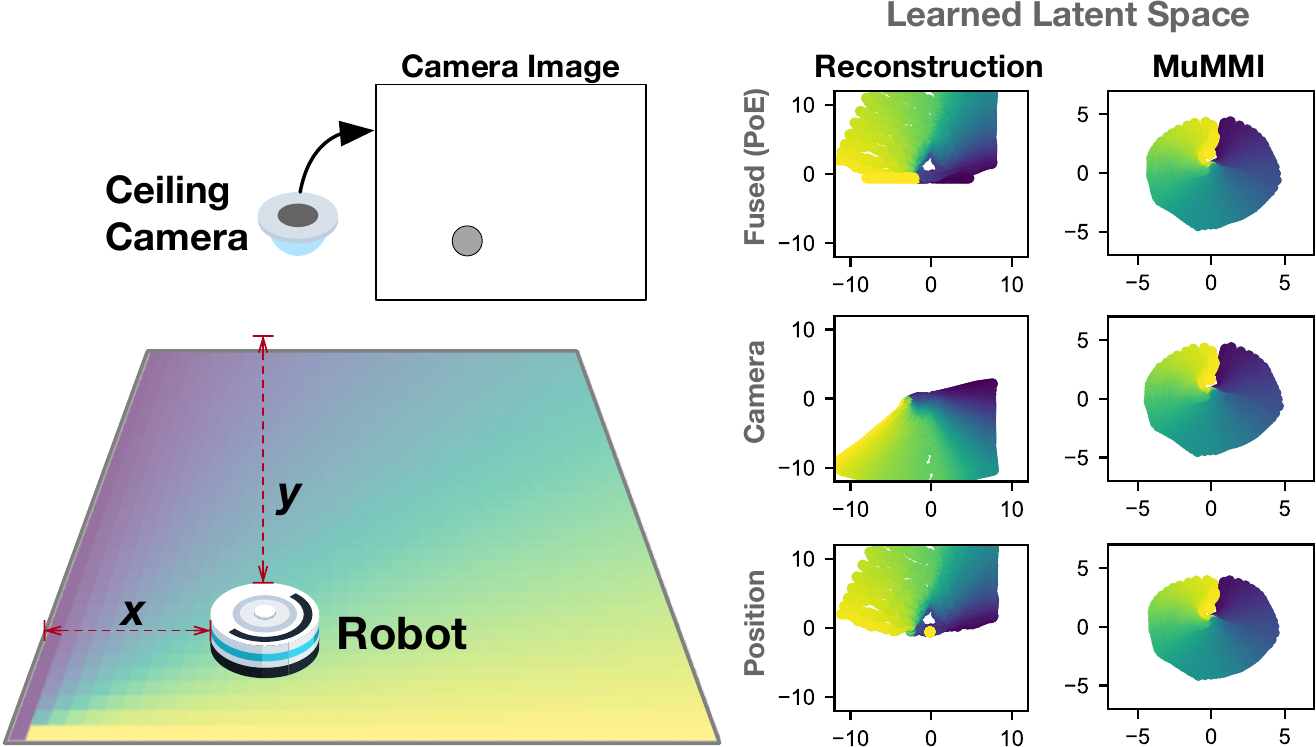}
    \caption{\small A simple illustrative example of a robot in a 2D world with two sensory modalities: laser rangers that give its $(x,y)$ position and a ceiling camera provides a scene image. A Deep Latent Space Model (SSM) trained using Product-of-Experts (PoE) fusion and a reconstruction-based loss did \emph{not} learn a robust latent space from gathered data (left plots, colors indicate ground-truth position): the overlap between the two modality-specific latent spaces is small and model is over-reliant on the position sensor. The experts were ``miscalibrated'' in that the camera expert predicts a much higher variance relative to the position expert and thus, has little influence during PoE fusion. In contrast, our proposed MuMMI training encourages a consistent latent space across the different modalities (right plots) with calibrated experts.}
    \label{fig:simple_example}
\end{figure}

Our key contribution is a mutual-information (MI) driven training method. 
Prior works have trained multimodal deep models by maximizing a reconstruction-based variational evidence lower-bound (ELBO) of the log data likelihood~\cite{wu2018mvae,zhi2020factorized,hafner2019dream}. Our insight is that the standard ELBO does not sufficiently enforce a \emph{shared} latent space between the different modalities. As a result, the learned world models do not well-integrate information from multiple sensors and the learned space is poorly structured (see Fig. \ref{fig:simple_example} and additional plots in the online appendix~\cite{chen2021app}). As a remedy, we derive a MI-based lower-bound that is optimized via the InfoNCE loss~\cite{oord2018infoNCE}. Within this contrastive framework, we explicitly encourage the different modality networks to be consistent with one another via a specially-designed density ratio estimator. Unlike prior work on self-supervised RL with multiple modalities~\cite{lee2019feifeili1,lee2020feifeili2}, our methodology is task-independent and alleviates the need to craft task/sensor-specific semi-supervised losses. 

Experiments show that our Multi-Modal Mutual Information (MuMMI) approach significantly outperforms existing state-of-the-art techniques for self-supervised RL~\cite{hafner2019dream,ma2020contrastive} on Natural MuJoCo tasks~\cite{ma2020contrastive} augmented with additional modalities. A further preliminary experiment on the challenging Robosuite Table Wiping task~\cite{robosuite2020} shows that MuMMI is able to learn policies that are robust to a missing sensor. Specifically, inputs from two RGB cameras (one workspace camera and another mounted on the robot) were provided during training. During testing, we observed policy performance remained comparable even when completely removing the workspace camera.

In summary, this paper presents three key contributions:
\begin{itemize}
	\item The Multi-Modal State-space model (MSSM), which can represent complex dynamics and multi-modal observations;
	\item The MuMMI training loss that encourages modalities to share a common latent space, which promotes  robustness to missing observations;
	\item Empirical results showing that the MSSM trained with MuMMI outperforms competing methods and ablated variants, which indicate the importance of a modular structure and a shared latent space.
\end{itemize}

\section{Preliminaries: Latent State-Space Models}
\label{sec:background}

\begin{figure*}
    \centering
    \includegraphics[width=0.80\textwidth]{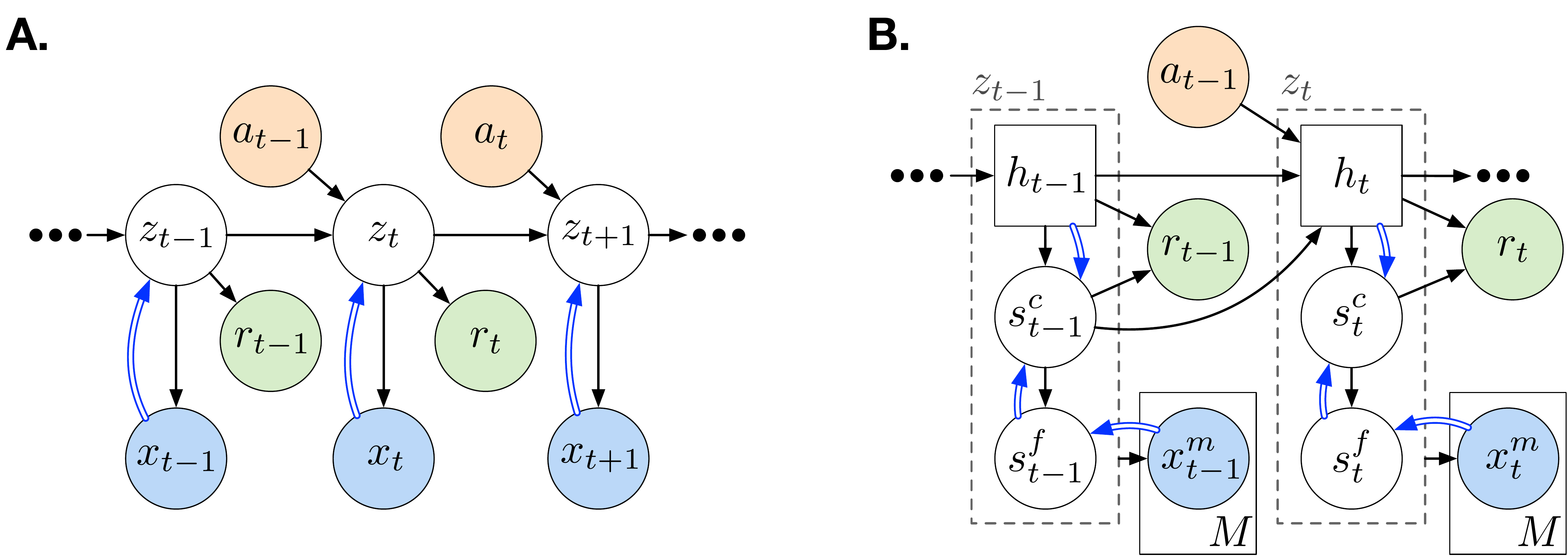}
    \caption{\small Probabilistic Graphical Models (PGMs) for (\textbf{A.}) the basic latent state-space model (SSM) used in reinforcement learning/planning contexts and (\textbf{B.}) our Multi-Modal State-Space Model (MSSM). Circle nodes represent random variables and shaded nodes are observed. The square nodes indicate a deterministic function mapping.
    The MSSM generalizes the basic model to $M$ modalities and decomposes the latent variables $z_t = [h_t, s^c_t, s^f_t]$. Inference networks $q_\phi$ are shown using blue double-lined arrows; we see that $q(s^f_t|x_t^{1:M})$ fuses information across the modalities, whilst $q(s_{t}^{c}|s_{t}^{f},h_{t})$ encodes information from the latent dynamics and the fused observations. Please see main text in Secs. \ref{sec:background} and \ref{sec:modelstructure} for additional details.}
    \label{fig:pgms}
    \vspace{-1em}
\end{figure*}

Latent state-space models (SSMs) have been a long-standing staple of robotics. For example, the popular Kalman filter~\cite{kalman1960new} comprises Gaussian latent (hidden) random variables with linear transitions between time-steps and linear observation functions. Other example SSMs include Hidden Markov Models~\cite{rabiner1986introduction} for discrete latent spaces, and probabilistic SLAM models~\cite{thrun05probrob}. This section assumes familiarity with probabilistic graphical models (PGMs); please refer to \cite{jordan2003introduction} for an excellent introduction.  

Modern-day SSMs that leverage deep neural networks are able to capture complex nonlinear transitions and rich high-dimensional observations (e.g., camera images). Figure \ref{fig:pgms}.A. illustrates a prototypical SSM where the $z_t$'s are latent states from which the observations $x_t$'s are generated. Transitions between time-steps $t$ are Markovian and conditioned upon actions $a_t$ taken by the robot. In reinforcement learning (RL) settings, we also include a reward per time-step $r_t$; here, we consider state-dependent reward distributions. Given the probabilistic graphical model in Fig. \ref{fig:pgms}.A., the joint distribution of the model factorizes as:
\begin{align}
    p_\theta(&x_{1:T},r_{1:T},z_{1:T}|a_{1:T}) = \nonumber \\
    & \prod_{t=1}^{T}p_\theta(x_{t}|z_{t})p_\theta(r_{t}|z_{t})p_\theta(z_{t}|z_{t-1},a_{t-1})
    \label{eqn:basicfactorization}
\end{align}
where $\theta$ are model parameters, $x_{1:T}$ denotes all observations from $t=1,\dots,T$, and likewise for $r_{1:T}$ ,$z_{1:T}$ and $a_{1:T}$. The three distributions in the factorization above correspond to:
\begin{align}
    \mathrm{\textrm{Observations:} } & \quad p_\theta(x_{t}|z_{t}) \\
    \mathrm{\textrm{Rewards:} } & \quad p_\theta(r_{t}|z_{t}) \\
    \mathrm{\textrm{Transitions:} } & \quad p_\theta(z_{t}|z_{t-1},a_{t-1})& 
\end{align}
and can be modelled using nonlinear function approximators such as deep neural networks.

One can view the model above as a Partially-Observable Markov Decision Process (POMDP)~\cite{astrom1965pomdpl,kaelbling1998pomdp} that is specified up to the unknown parameters $\theta$. We would like to learn $\theta$ from observed data, $\mathcal{D} = \left\{x_{t}, a_{t}, r_{t}\right\}_{t=1}^{T}$, but maximum likelihood estimation is generally intractable as we need to marginalize out the latent $z_t$'s. As such, we optimize the evidence lower bound (ELBO) under the data distribution $p_d$, i.e., $\mathbb{E}_{p_d}[\mathcal{L}_e] \leq \mathbb{E}_{p_d}[\log   p_\theta(x_{1:T},r_{1:T}|a_{1:T})]$, where 
\begin{align}
\label{eqn:ELBO}
     \mathcal{L}_e = & \sum_{t=1}^{T}\Big(\displaystyle\mathop{\mathbb{E}}_{q_{\phi}(z_{t})}\left[\log p_{\theta}(x_{t}|z_{t})\right] +\displaystyle\mathop{\mathbb{E}}_{q_{\phi}(z_{t})}\left[\log p_{\theta}(r_{t}|z_{t})\right]  \nonumber\\
    &- \displaystyle\mathop{\mathbb{E}}_{q_{\phi}(z_{t-1})}\left[\DKL\left[q_{\phi}(z_{t}) \| p_{\theta}(z_{t}|z_{t-1},a_{t-1})\right]\right]\Big)
\end{align}
using a variational distribution $q_\phi$, which is typically an \emph{inference network} parameterized by $\phi$. For simplicity, we denote the inference network as $q_\phi(z_t)$, but keep in mind the distribution is often conditioned on observations, e.g., $q_\phi(z_t|x_t)$.
In the ELBO, the first two reconstruction terms encourage encoding of information of $x_t$ and $r_t$ in the latent state $z_t$. The third Kullback-Leibler (KL) divergence term enforces consistency between the variational distribution  $q_\phi$ and the transition dynamics $p_{\theta}(z_{t}|z_{t-1},a_{t-1})$. 

\section{Multi-Modal Deep Latent State-Space Model}
\label{sec:modelstructure}

In this section, we describe our Multi-modal state-space model (MSSM), which extends the aforementioned SSM to multiple sensory modalities. We first describe the model structure, and then proceed to detail our MI-based training methodology. 

\para{Model Structure.} As a guide, Fig. \ref{fig:pgms}.B. illustrates a two-time-slice view of our model. Compared to the vanilla variant in Fig. \ref{fig:pgms}.A., MSSM generates multiple observations corresponding to the $M$ different modalities ($x^m_{t}$ in the plates) and employs a modified Recurrent SSM (RSSM) structure~\cite{hafner2019Planet} --- we decompose the latent state $z_{t}$ into three variables $z_{t}=\left[h_{t}, s^c_t, s^f_t\right]$. This splits the latent state into deterministic and stochastic parts; the transition governing $h_t$ is deterministic, which helps the model better remember previous states.
Unlike prior work~\cite{hafner2019Planet}, we further decompose the stochastic variable: $s^f_t$ encodes information about the current observations across modalities, whilst the ``combined'' stochastic variable $s^c_t$ also encodes past information. We find that this decomposition, when combined with appropriate inference networks, enables faster and more stable training. The joint distribution of the model factorizes in a similar manner to eq. (\ref{eqn:basicfactorization}):
\begin{align}
    p_\theta(x_{1:T}^{1:M},&r_{1:T},z_{1:T}|a_{1:T})  = \nonumber \\
    & \prod_{t=1}^{T}\prod_{m=1}^{M}p_\theta(x_{t}^{m}|z_{t})p_\theta(r_{t}|z_{t})p_\theta(z_{t}|z_{t-1},a_{t-1})
\end{align}
where $x_{1:T}^{1:M}$ denotes all the $M$ observations at every time step $t=1,\dots,T$, and 
\begin{align}
    \mathrm{\textrm{Observations:} } \quad & p_\theta(x_{t}^{m}|z_{t}) =p_\theta(x_{t}^{m}|s^{c}_{t}, h_{t}) \\
    \mathrm{\textrm{Rewards:} } \quad & p_\theta(r_{t}|z_{t}) =p_\theta(r_{t}|s^{c}_{t}, h_{t}) \\
    \mathrm{\textrm{Transitions:} } \quad & p_\theta(z_{t}|z_{t-1},a_{t-1}) = p_\theta(s^{f}_{t}|s^{c}_{t}) \nonumber\\ &\quad\quad p_\theta(s^{c}_{t}|h_{t})g_\theta(h_{t}|s_{t-1}^{c},h_{t-1},a_{t-1}) 
\end{align}
Note that the function $g$ above is deterministic (indicated by squares in Fig. \ref{fig:pgms}.B.).

\para{Model Training via Standard ELBO.} As in the single modality case, a possible training option is to maximize the ELBO,
\begin{align}
     \log &\, p_\theta(x_{1:T}^{1:M},r_{1:T}|a_{1:T}) \geq \mathcal{L}^M_e\nonumber \\ 
     =&\sum_{t=1}^{T} \Big( \sum_{m=1}^{M}{\mathbb{E}}_{q_{\phi}(z_{t})}\left[\log p_{\theta}(x_{t}^{m}|z_{t})\right]+ {\mathbb{E}}_{q_{\phi}(z_{t})}\left[\log p_{\theta}(r_{t}|z_{t})\right]  \nonumber\\
    &- {\mathbb{E}}_{q_{\phi}(z_{t-1})}\left[\DKL\left[q_{\phi}(z_{t})\|p_{\theta}(z_{t}|z_{t-1},a_{t-1})\right]\right]\Big)
\label{eqn:MELBO}
\end{align}
using the variational distribution,
\begin{align}
 &q_{\phi}(z_{1:T}|x_{1:T}^{1:M},a_{1:T})=\prod_{t=1}^{T}q(z_{t}|z_{t-1},x_{t}^{1:M}, a_{t-1})\nonumber\\
 &=\prod_{t=1}^{T}\left[\prod_{m=1}^{M}q(s_{t}^{f}|x_{t}^{m})\right]q(s_{t}^{c}|s_{t}^{f},h_{t})g_\theta(h_{t}|s_{t-1}^{c},h_{t-1},a_{t-1})    
\label{eqn:post_fact}
\end{align}
where $q(s_{t}^{f}|x_{t}^{m})$, $q(s_{t}^{c}|s_{t}^{f},h_{t})$ and $p(s_{t}^{c}|h_{t})$ are Gaussians, and the different modalities are fused via a Product-of-Experts (PoE)~\cite{hinton2002training}. If modality $m$ is missing, we can simply drop corresponding expert $q(s_{t}^{f}|x_{t}^{m})$. 

One key problem with maximizing the ELBO above is that the objective is under-constrained: the different modality experts need not share the same latent space. Prior work has primarily resorted to randomly dropping modalities during training to force consistency, but our experiments showed this approach may not be robust (Fig. \ref{fig:simple_example}). 

\para{Model Training via MuMMI.} In this work, we pursue an alternative information-theoretic approach, which turns out to be equivalent to maximizing $\mathcal{L}^M_e$ under specific assumptions. Let us define $v^{\setminus m}_{1:T} = (x_{1:T}^{\setminus m}, z_{1:T})$ where $x_{1:T}^{\setminus m}$ denotes observations from all the modalities \emph{except} modality $m$. To reduce clutter, we will drop the explicit dependence on $a_{1:T}$. Assume that the data is generated from the MSSM and consider the mutual information between $x^m_{1:T}$ and $v^{\setminus m}_{1:T}$:
\begin{align}
    \mathbb{I}[x^m_{1:T}&; v^{\setminus m}_{1:T}] = \sum p(x^m_{1:T}, v^{\setminus m}_{1:T}) \log \frac{ p(x^m_{1:T}, v^{\setminus m}_{1:T}) }{p(x^m_{1:T})p(v^{\setminus m}_{1:T}) } \nonumber \\
    &= \sum p(x^{1:M}_{1:T})p(z_{1:T}|x^{1:M}_{1:T}) \log \frac{ p(x^m_{1:T} | z_{1:T}) }{p(x^m_{1:T}) } \nonumber \\
    &=  {\mathbb{E}}_{p(x^{1:M}_{1:T})p(z_{1:T}|x^{1:M}_{1:T})}\left[ \log p(x^m_{1:T} | z_{1:T}) \right] - C^m
    \label{eqn:infoterm}
\end{align}
where $C^m = {\mathbb{E}}_{p(x^m_{1:T})}\left[p(x^m_{1:T})
\right]$, and we have leveraged the conditional independence assumptions in the MSSM when dropping the dependence on $x_{1:T}^{\setminus m}$ in $p(x^m_{1:T} | x_{1:T}^{\setminus m}, z_{1:T})$. Intuitively, $\mathbb{I}[x^m_{1:T}; v^{\setminus m}_{1:T}]$ captures the mutual dependence between a given modality $m$ and the remaining observations together with the latent state. 
If we assume that $q(z_{1:T}|x^m_{1:T}) = p(z_{1:T}|x^{1:M}_{1:T})$, we can combine eq.~(\ref{eqn:MELBO}) and eq.~(\ref{eqn:infoterm}) to yield
\begin{align}
    \mathbb{E}_{p_d}&[\mathcal{L}_e^M] \nonumber \\
    & = \sum_{m=1}^M (\mathbb{I}[x^m_{1:T}; v^{\setminus m}_{1:T}] + C^m) + {\mathbb{E}}_{p_dq_{\phi}(z_{t})}\left[\log p_{\theta}(r_{t}|z_{t})\right]   \nonumber\\ 
    & \quad - {\mathbb{E}}_{p_d q_{\phi}(z_{t-1})}\left[\DKL\left[q_{\phi}(z_{t})\|p_{\theta}(z_{t}|z_{t-1},a_{t-1})\right]\right]
\end{align}
which relates $\mathbb{I}[x^m_{1:T}; v^{\setminus m}_{1:T}]$ to the ELBO. For the purposes of learning,  $\sum_m C^m$ is a constant that does not depend on the parameters $\theta$, and can be dropped. 

To optimize $\mathbb{I}[x^m_{1:T}; v^{\setminus m}_{1:T}]$, we use the InfoNCE loss~\cite{oord2018infoNCE}. Let us define the density ratio estimator,
\begin{align}
	f^m_\theta(x^m_{1:T}, v^{\setminus m}_{1:T}) \propto \frac{p(x^m_{1:T} | v^{\setminus m}_{1:T})}{p(x^m_{1:T})} = \frac{p(x^m_{1:T} | z_{1:T})}{p(x^m_{1:T})}
	\label{eqn:fdrest}
\end{align}
where we have again exploited the conditional independence between modalities given $z_{1:T}$. As such, we can specify $M$ density ratio estimators independently for each modality, which simplifies our setup and eases computational burden. Abusing notation, we let $f^m_\theta(x^m_{1:T}, z_{1:T}) = f^m_\theta(x^m_{1:T}, v^{\setminus m}_{1:T})$. From \cite{oord2018infoNCE}, we can show that,
\begin{equation}
	\mathbb{I}[x^m_{1:T}; v^{\setminus m}_{1:T}] \geq \mathbb{E}\left[\log\frac{f^m_\theta(x^{+,m}_{1:T}, z_{1:T})}{\sum_{x^{-,m}_{1:T}} f^m_\theta(x^{-,m}_{1:T}, z_{1:T})}\right]
	\label{eq:infoncebound}
\end{equation}  
where $x^{+,m}_{1:T}$ and $x^{-,m}_{1:T}$ are ``positive'' and ``negative'' samples, respectively.  We obtain the positive sample by drawing $x^{+,m}_{1:T} \sim p_\theta(x^m_{1:T} | z_{1:T})$ and $N-1$ negative samples from the proposal distribution $p_d(x^m_{1:T})$.

\begin{figure}
    \centering
    \includegraphics[width=0.8\columnwidth]{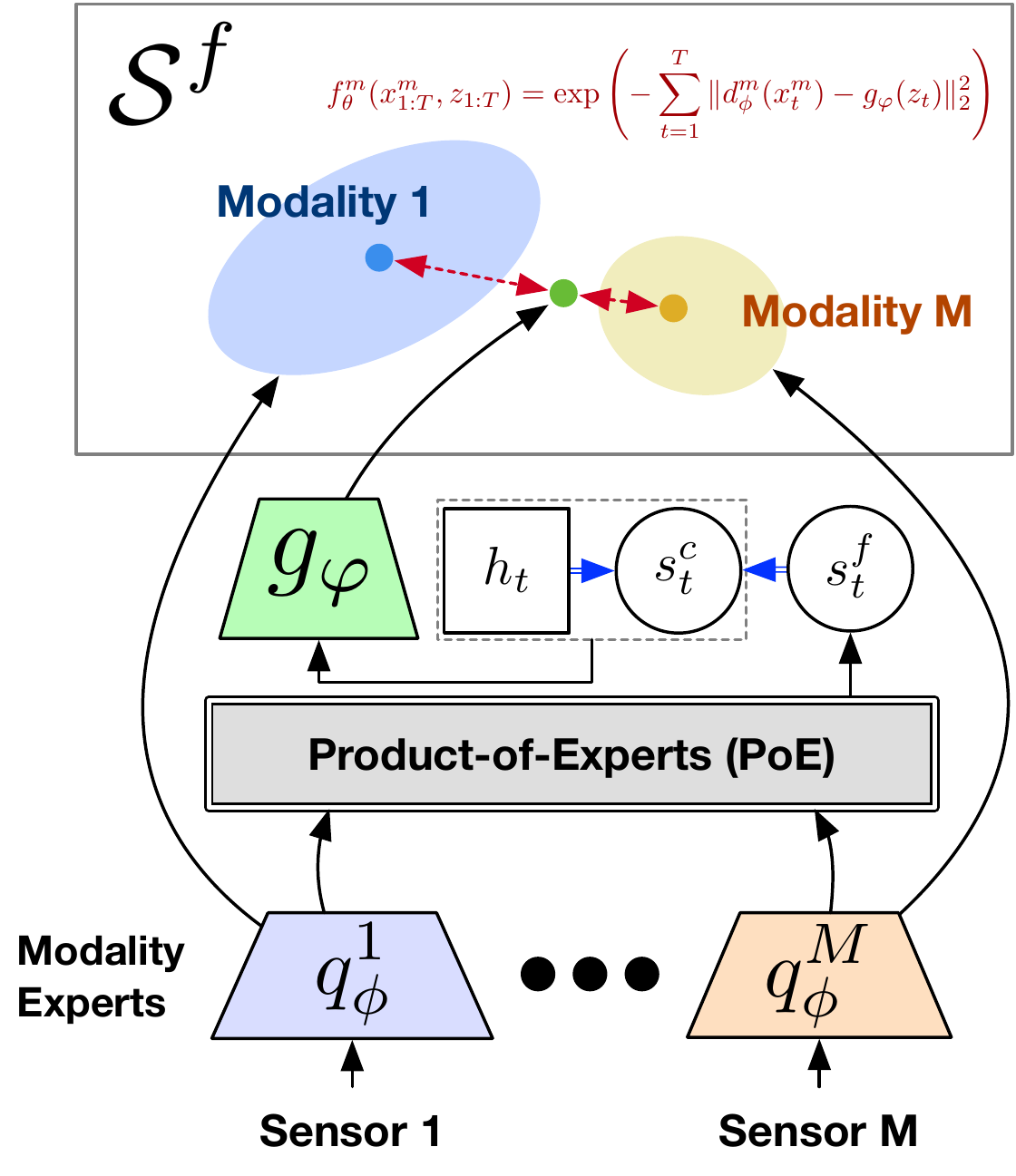}
    \caption{\small MuMMI training uses a density ratio estimator $f^m_\theta$ (eq.~(\ref{eqn:fdrest})) that acts to minimize the squared distances between the mean of each modality expert and a transformed fused latent code. This encourages the experts to project to points in a shared latent space.}
    \label{fig:MuMMI}
\end{figure}

Although the InfoNCE is a looser bound of the log marginal likelihood, it affords us additional design freedom in the density ratio estimator $f^m_\theta$. 
We propose a design that encourages \emph{each} modality expert to map data to points close to a (fused) latent state:
\begin{align}
    f_{\theta}^{m}(x_{1:T}^{m}, z_{1:T}) = \exp& \left(-\sum_{t=1}^{T}\| d_\phi^{m}(x_{t}^{m})- g_{\varphi}(z_{t})\|_2^2\right)
\label{eqn:kernel_func}
\end{align}
where $d_{\phi}^{m}$ and $g_{\varphi}$ are also neural networks. 
We set $d_{\phi}^{m}$ to share parameters with $q_{\phi}(s_{t}^{f}|x_{t}^{m}) = \mathcal{N}(\mu^m_\phi(x^m_{t}), v^m_\phi(x^m_{t}))$, i.e., $d_{\phi}^{m} = \mu^m_\phi$. 
Here, $f$ can be seen as a squared exponential kernel and maximizing the numerator in eq. (\ref{eq:infoncebound}) across modalities encourages consistent projections (see Fig. \ref{fig:MuMMI}). 

\para{Final Loss and In-Practice.} Training the MSSM via MuMMI entails optimizing $\mathbb{E}_{p_d}[\widehat{\mathcal{L}}]$ where:
\begin{align}
\mathbb{E}_{p_d}[\widehat{\mathcal{L}}] & = \sum_{m=1}^M \lambda^m \mathbb{E}\left[\log\frac{f^m_\theta(x^{+,m}_{1:T}, z_{1:T})}{\sum_{x^{-,m}_{1:T}} f^m_\theta(x^{-,m}_{1:T}, z_{1:T})}\right]  \\
     & \quad+ {\mathbb{E}}_{p_d q_{\phi}(z_{t})}\left[\log p_{\theta}(r_{t}|z_{t})\right]   \nonumber\\ 
     & \quad- {\mathbb{E}}_{p_d q_{\phi}(z_{t-1})}\left[\DKL\left[q_{\phi}(z_{t})\|p_{\theta}(z_{t}|z_{t-1},a_{t-1})\right]\right]\nonumber
\end{align}
and the $\lambda^m$'s are hyperparameters, which can be set equal or tuned using prior knowledge of which modality is more informative. To compute $f^m_\theta$, we use a strategy similar to prior work~\cite{ma2020contrastive}: we sample a batch of sequences $\{x_{1:T}^{1:M,i},a_{1:T}^{i},r_{1:T}^{i}\}_{i=1}^{B}$ from a replay buffer, where $B$ is the batch size. For each state-observation pair, we treat the other $(B\times T)-1$ observations in the same batch as negative samples.

\section{Related Work}
\label{sec:related_work}

Our work builds upon recent advances in probabilistic multi-modal models and deep reinforcement learning. Specifically, MuMMI uses PoE fusion~\cite{hinton2002training}, which was previously used in a multi-modal variational autoencoder~\cite{wu2018mvae} that was later extended to sequential settings~\cite{zhi2020factorized}. Multi-modal models have also been adopted in robotics applications, where feature vectors from different modalities are concatenated into a single latent representation~\cite{zambelli2020multimodal,lee2019feifeili1}. Lately, PoE-based fusion has been applied to multi-modal self-supervised training~\cite{lee2020feifeili2}, but unlike MuMMI, the method relies on hand-crafted task-dependent losses. In a related research thread, very recent work has explored event-driven multi-modal representations using Spiking Neural Networks~\cite{taunyazov20event}. Here, we use deep artificial neural networks but MuMMI can potentially be extended to event-driven learning. 

MuMMI is also related to recent self-supervised model-based RL methods, e.g., PlaNET~\cite{hafner2019Planet} and Dreamer~\cite{hafner2019dream}, which learn latent dynamics models via interactions with the environment. The backbone of these methods is the RSSM model, on which our MSSM is based. However, these techniques rely the standard reconstruction-based ELBO, which is not robust to irrelevant noise. 
Our approach is closely related to the recently proposed CVRL~\cite{ma2020contrastive}, which learns using the InfoNCE loss. However, CVRL (and other self-supervised RL methods) have largely focused on single-modality learning with reliable sensors. Unlike the works above, MuMMI trains a multi-modal world model (the MSSM) that is demonstrably robust to missing data.

\section{Experiment: Multi-Modal Natural MuJoCo }
\label{sec:selfsupRL}

\begin{figure}
  \centering
  \includegraphics[width=0.9\columnwidth]{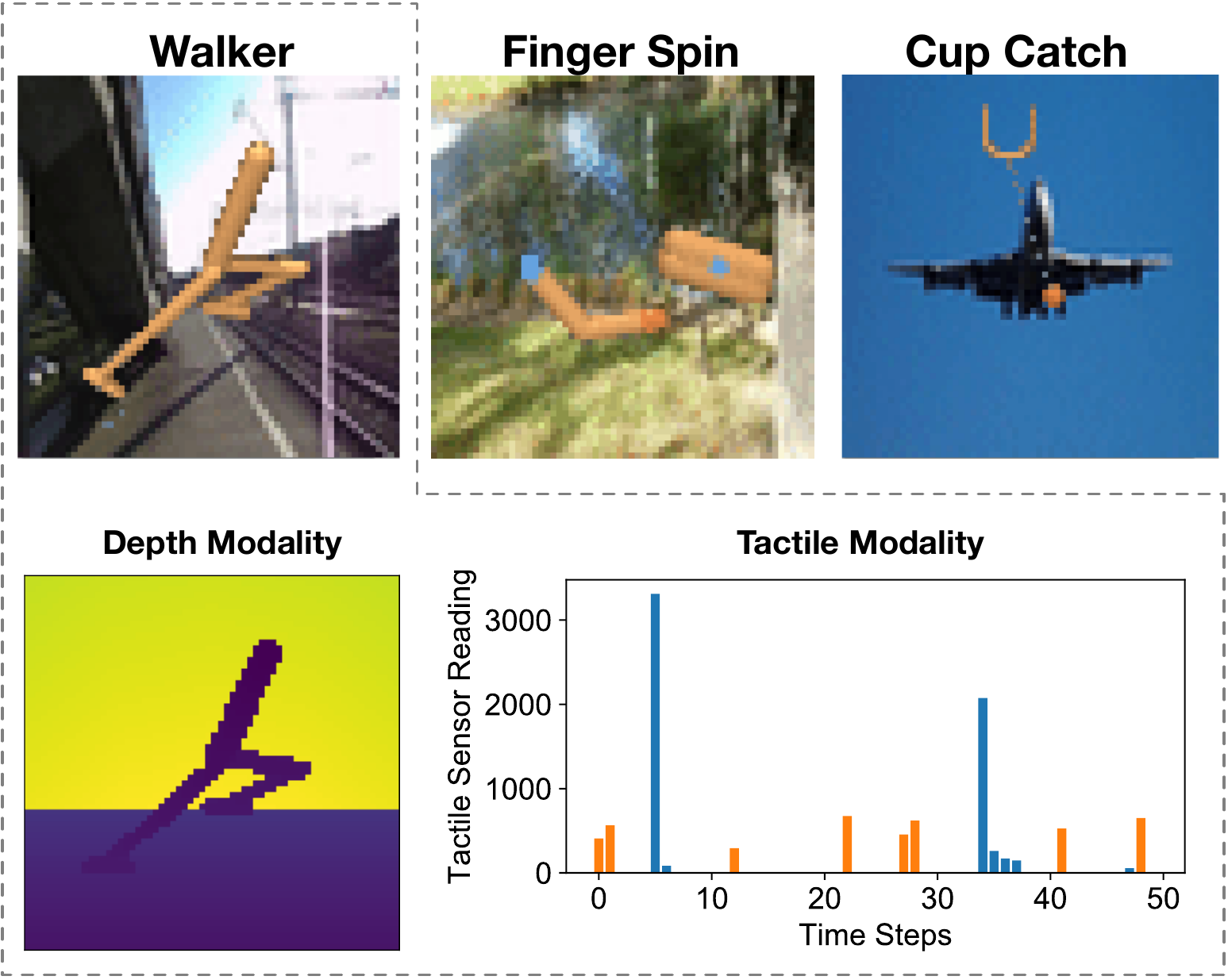}
  \caption{\small Natural Mujuco environments and modalities used in our experiments (\textbf{top}) RGB images for walker stand/walk/run, finger spin and cup catch. The background images are continuously changing. (\textbf{bottom}) Additional two modalities for the walker, i.e., depth image and tactile sensor. The tactile sensors were placed on the soles of the feet and activated when it came into contact with the ground. For the other two environments, the tactile sensor was placed on the finger-tip and inside the cup, respectively.}
  \label{fig:exp_modalities}
  \vspace{-0.5cm}
\end{figure}

\begin{figure*}
    \centering
    \includegraphics[width=0.95\textwidth]{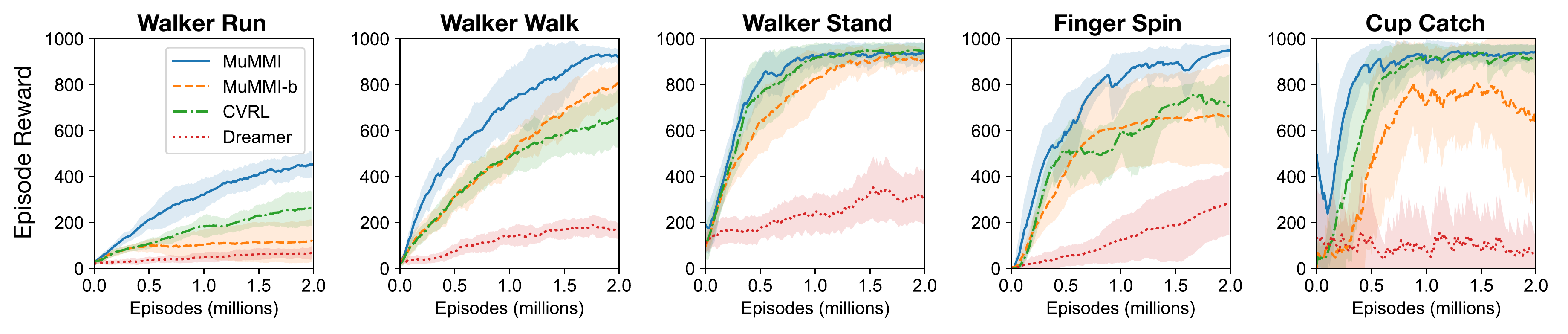}
    \caption{\small Model performance across training episodes (moving average smoothing with weight 0.9). The performance curves are steeper for MuMMI on a majority of the tasks, indicating faster learning compared to  MuMMI-b and competing approaches (Dreamer and CVRL).}
    \label{fig:natmujocores}
     \vspace{-0.1cm}
\end{figure*}

\begin{table*}
\centering
\caption{\small Multi-Modal Natural Mujoco Experiment. Performance measured by Mean Total Episode Reward (averaged over 30 episodes, with standard deviations). Highest average reward in \textbf{bold}.}
\label{tbl:RLResults}
\small  
\begin{tabular}{c c c c c c }
\hline
\hline 
    Task  &  Missing Data & MuMMI & MuMMI-b & CVRL & Dreamer \\
\hline
     ~ & None              & $\textbf{466.7}\pm25.5$ & $116.4\pm13.6$ & $311.7\pm30.2$ & $71.3\pm12.4$ \\ 

     walker run & Medium   & $\textbf{438.7}\pm33.1$ & $119.1\pm12.1$ & $277.5\pm22.4$ & $70.7\pm9.70$ \\ 
    ~ & High               & $\textbf{382.4}\pm38.5$ & $105.2\pm14.6$ & $255.8\pm31.4$ & $70.9\pm37.3$ \\ 
\hline
    ~ & None               & $\textbf{954.5}\pm21.5$ & $870.4\pm44.1$ & $732.7\pm61.7$ & $182.9\pm37.7$ \\ 

     walker walk & Medium  & $\textbf{939.9}\pm26.1$ & $813.1\pm65.1$ & $703.4\pm48.9$ & $163.1\pm31.8$ \\ 
     ~ & High              & $\textbf{890.0}\pm50.1$ & $760.3\pm61.5$ & $607.5\pm58.3$ & $172.8\pm39.2$ \\ 
\hline
     ~ & None              & $\textbf{966.8}\pm24.6$ & $959.8\pm34.5$ & $955.6\pm27.6$ & $307.0\pm54.40$ \\ 

     walker stand & Medium & $\textbf{952.4}\pm28.7$ & $946.5\pm24.1$ & $940.1\pm34.3$ & $334.3\pm122.6$ \\ 
     ~ & High              & $\textbf{922.0}\pm59.7$ & $918.2\pm58.4$ & $918.2\pm31.9$ & $280.4\pm72.70$ \\ 
\hline
    ~ & None               & $\textbf{965.6}\pm10.9$ & $679.1\pm38.7$ & $705.9\pm32.1$ & $342.7\pm37.3$ \\ 

    finger spin & Medium   & $\textbf{956.7}\pm17.0$ & $661.4\pm42.7$ & $685.2\pm36.4$ & $303.6\pm30.7$\\ 
    ~ & High               & $\textbf{929.1}\pm20.8$ & $620.4\pm58.2$ & $623.8\pm39.1$ & $271.2\pm35.1$\\ 
\hline
    ~ & None               & $\textbf{949.5}\pm34.3$ & $725.8\pm249.2$ & $922.0\pm48.1$ & $124.0\pm290.3$ \\ 

    cup catch & Medium     & $\textbf{948.0}\pm32.9$ & $637.8\pm335.3$ & $904.2\pm78.7$ & $47.7\pm147.3$ \\ 
    ~ & High               & $\textbf{927.3}\pm38.0$ & $540.5\pm317.7$ & $922.3\pm47.9$ & $73.5\pm225.3$ \\ 
\hline
\hline
\end{tabular}
\end{table*}

In this section, we describe experiments designed to evaluate the MSSM and MuMMI on the task of self-supervised RL; for simplicity, we will refer to the MSSM with MuMMI training as MuMMI. Our goal was to ascertain whether MuMMI led to better performance and robustness to missing data, compared to competing state-of-the-art methods. 

\para{Methods. } We compare MuMMI against two representative state-of-the-art model-based deep RL methods: Dreamer~\cite{hafner2019dream} and CVRL~\cite{ma2020contrastive}. Dreamer uses a reconstruction-based ELBO, whilst CVRL is trained using a contrastive loss (but without a product-of-experts fusion layer). For both models, feature vectors extracted from modality-specific deep networks are fused via concatenation and missing observations are masked with zeros (similar to prior work~\cite{zhi2020factorized, zambelli2020multimodal}). We also tested MuMMI-b; a variant of MuMMI with a modified density ratio estimator: $f_{b}(x_{1:T}^{1:M}, z_{1:T}) = \exp \left(-\sum_{t=1}^{T}\| b_\phi(x_{t}^{1:M})-g_{\varphi}(z_{t})\|_2^2\right)$, where $b_\phi(x_{t}^{1:M})$ is set to the mean of the fused PoE distribution. 
Compared to eq. (\ref{eqn:kernel_func}), $f_{b}$ promotes consistency between the PoE-fused latent vectors and the learned dynamics. It does not directly constrain \emph{individual} modalities have similar latent codes, but may work well if given sufficient data (and trained using random drops~\cite{wu2018mvae}). 
All methods used latent imagination, an actor-critic RL method~\cite{hafner2019dream} and latent-guided MPC~\cite{ma2020contrastive}. 

\para{Multi-Modal Tasks. } We used the MuJoCo-powered DeepMind Control Suite~\cite{tassa2020dmcontrol}, but augmented to have complex backgrounds (Natural MuJoCo~\cite{ma2020contrastive}) and additional modalities (Fig. \ref{fig:exp_modalities}). The standard benchmarks already pose challenges common to robot learning: sparse rewards, high-dimensional 3D scenes, many degrees of freedom, and contact dynamics. The complex backgrounds---videos from ILSVRC dataset~\cite{russakovsky2015imagenet}---add a degree of realism and difficulty as the robot needs to separate useful information from irrelevant noise. We selected 5 benchmark tasks based on available computational budget. The modalities for all tasks comprise RGB and depth images, and tactile feedback. The backgrounds are assumed far and do not appear in the depth images; this tests if the models are able to use this ``clean'' modality to improve performance, yet not become overly reliant on it. The tactile modality has significantly different properties compared to the images; it is a sparse signal that occurs when certain parts (i.e., the walker feet, finger tip, and inside-cup) come into contact with the ground or other objects.  

\para{Methodology. } 
For each task-method pair, we conducted 3 training sessions where each session was initialized with a different random seed and trained for 2 million episodes. Each session took $\approx$ 1 day to complete on a workstation with a Nvidia 2080Ti GPU.  
During training, data was randomly dropped to simulate data loss (e.g., from faulty sensors or occlusions); for each modality, we dropped segments of varying lengths (the start and length of missing segments are uniformly random, but constrained so that the missing rate was $37.5\%$ of the complete data).
In the testing stage, we compared each method's accumulated rewards per-episode (averaged over the 3 trained policies). Each policy was tested over 3 batches of 10 episodes, where each batch with a different missing rate (None: $0\%$; Medium: $37.5\%$; High: $75\%$). Complete model architecture details and source code is available in the online appendix.

\begin{table*}[hbt!]
\centering

\caption{\small Table Wiping Performance measured by Mean Total Episodic Reward (averaged over 30 episodes, with standard deviations)}.
\label{tbl:CSResults}
\small  
\begin{tabular}{c c c c }
\hline
\hline 
    Modalities & MuMMI & MuMMI-b & MSSM-e \\
\hline
    All Modalities (Medium Missing Data)  & $61.4\pm70.4$ & $48.8\pm40.5$ & $60.9\pm56.1$ \\
    
    All Modalities (Full Observed)  & $60.6\pm69.9$ & $47.9\pm37.2$ & $64.9\pm85.3$ \\ 
    
    Robot Camera (Medium Missing Data) & $57.8\pm69.8$ & $62.1\pm93.7$ & $64.0\pm95.9$ \\ 

    Robot Camera (Full Observed) & $65.9\pm76.9$ & $69.2\pm97.2$ & $60.4\pm98.2$ \\ 
\hline
\hline
\end{tabular}
    \vspace{-0.3cm}
\end{table*}

\para{Results.} The final performance of the different models is summarized in Table \ref{tbl:RLResults}. On all of the tasks, MuMMI outperforms all other competing approaches by a significant margin. The poorer performance of the ablated MuMMI-ab indicates the importance of a common latent space for PoE fusion. We observed that MuMMI degrades gracefully with greater amount of missing data, but remains robust compared to the other methods. Between the concatenation fusion methods (Dreamer and CVRL), Dreamer has poorer performance, despite given access to the clean depth images\footnote{Given a single modality of clean image data, Dreamer is generally able to achieve high rewards on the tasks tested~\cite{hafner2019dream,ma2020contrastive}.}. In comparison, CVRL was better able to learn from multiple modalities;  we posit that Dreamer reconstruction loss does not permit the model to neglect the irrelevant inputs, which hampered learning of a good latent code. Finally, we observed that MuMMI learns faster than other methods, as indicated by the steeper learning curves in Fig. \ref{fig:natmujocores}.

\section{Case Study: Table Wiping}
In this section, we describe preliminary experiments using MuMMI to train a Franka-Emika Panda arm on the challenging Table Wiping benchmark task~\cite{robosuite2020}. 
Due to space constraints, we describe the essentials; please see the online appendix for additional information. We compared three methods: MuMMI, MuMMI-b and MSSM-e. MSSM-e is trained using a reconstruction-based ELBO (similar to Dreamer), but uses PoE instead of concatenation to fuse the modalities. We trained MSSM-e using a similar approach as \cite{wu2018mvae} where missing input modalities are dropped.

\begin{figure}
    \centering
    \includegraphics[width=0.8\columnwidth]{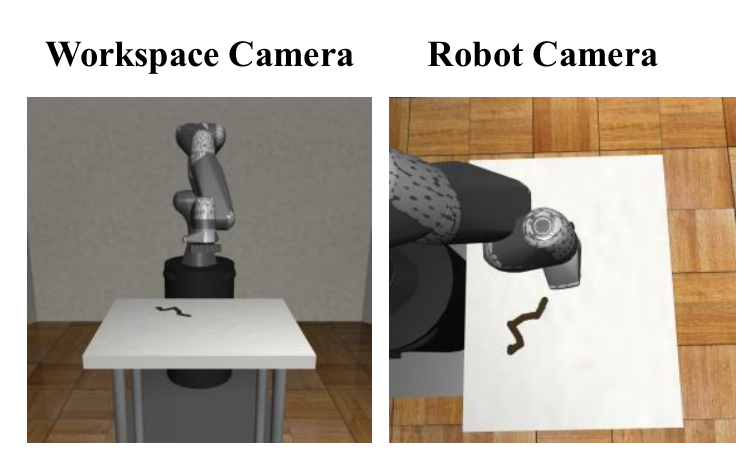}
    \caption{\small Views from the two RGB cameras used for Table Wiping. }
    \label{fig:expwipe}
    \vspace{-0.3cm}
\end{figure}

In the Table Wiping task, the Panda robot has to clean a table by erasing markings on its surface. The markings are randomized at the start of each episode. This task is one of the more challenging benchmarks in Robosuite and previous work using the state-of-the-art model-free soft-actor critic (SAC)~\cite{haarnoja2018soft} failed solve the task~\cite{robosuite2020}. Here, the robot can access two modalities: a RGB camera mounted on the top of the robot and a workspace RGB camera (Fig. \ref{fig:expwipe}).  

We trained each method for 1 million episodes with  domain randomization and moderate data loss ($37.5\%$) during training. In the testing stage, we compared each method's accumulated rewards per-episode (averaged over 30 episodes). Our results are summarized in Table \ref{tbl:CSResults}. We see that the methods were robust to removal of the workspace camera; performance was not drastically affected by the removal. Interestingly, we see that MSSM-e was also able to perform well for this particular problem. These preliminary results are promising; they show MuMMI and MSSM can be applied towards robotics problems in scenarios with unreliable sensors.
\section{Conclusions}
This work presents the MSSM and MuMMI. Together, they can be used to learn robust world-models from multi-modal sensory streams, even with significant amounts of missing data. 
Moving forward, we plan to apply MuMMI beyond self-supervised RL to other robot tasks including planning, human modeling, and imitation learning.

\section*{Acknowledgements}
This work was supported by the Science and Engineering Research Council, Agency of Science, Technology and Research, Singapore, through the National Robotics Program under Grant No. 192 25 00054.

\balance 
\bibliographystyle{IEEEtran}
\bibliography{references}

\clearpage
\section*{Appendix}

\subsection{Latent Imagination and Actor-Critic}
Following \cite{hafner2019dream}, after training, the agent generates the imagined trajectories using the learnt world model. Specifically, given a current state $z_{t}$, the agent sample the next imagined state by $\tilde{z}_{t+1}\sim p_{\theta}(z_{t+1}|z_{t-1},a_{t-1})$ and associated reward $\tilde{r}_{t+1}\sim p_{\theta}(r_{t+1}|z_{t+1})$ and the next action $\tilde{a}_{t}\sim \pi_{\eta}(a_{t+1}|z_{t+1})$. This process is repeated until an imagined trajectory $\left\{\tilde{z}_{t},\tilde{a}_{t},\tilde{r}_{t}\right\}_{t=\tau}^{\tau+H}$ is generated. Then, the agent learns the action and value models by optimizing:
\begin{equation}
\begin{split}
    &\max_{\eta}\displaystyle\mathop{\mathbb{E}}_{\pi_{\eta},p_{\theta}}\left(\sum_{t=\tau}^{\tau+H}V_{\lambda}(z_{t})\right)\\
    &\min_{\psi}\displaystyle\mathop{\mathbb{E}}_{\pi_{\eta},p_{\theta}}\left(\sum_{t=\tau}^{\tau+H}\frac{1}{2}\left\lVert v_{\psi}(z_{t})- V_{\lambda}(z_{t})\right\lVert^{2}\right)\\
\end{split}
\end{equation}

where
\begin{equation}
\begin{split}
    &V_{\lambda}(\tilde{z}_{t})=\displaystyle\mathop{\mathbb{E}}_{\pi_{\eta},p_{\theta}}\left(\sum_{n=\tau}^{h-1}\lambda_{n-\tau}\tilde{r}_{n}\right)\\
    &V_{\lambda}(\tilde{z}_{t})=(1-\lambda)\sum_{n=1}^{H-1}\lambda^{n-1}V_{N}^{n}(\tilde{z}_{t})+\lambda^{H-1}V_{N}^{H}(\tilde{z}_{t})\\
\end{split}
\end{equation}

The objective $\max_{\eta}\displaystyle\mathop{\mathbb{E}}_{\pi_{\eta},p_{\theta}}\left(\sum_{t=\tau}^{\tau+H}V_{\lambda}(z_{t})\right)$ optimizes the policy $\pi_{\eta}$ under current critic and the objective $\min_{\psi}\displaystyle\mathop{\mathbb{E}}_{\pi_{\eta},p_{\theta}}\left(\sum_{t=\tau}^{\tau+H}\frac{1}{2}\left\lVert v_{\psi}(z_{t})- V_{\lambda}(z_{t})\right\lVert^{2}\right)$ optimizes the value estimation. We also use latent guided model predictive control as in \cite{ma2020contrastive}. 

All in all, in each iteration, MuMMI first learns the world model  by using samples in replay buffers. Then, MuMMI use latent imagination to optimize the actor and critic. By iterating this process, the agent is able to learn behaviors in complex environments. To assist policy optimization, latent-guided MPC~\cite{ma2020contrastive} was also used.

\subsection{Multi-Modal State-Space Model}
We use the similar model architectures similar to  \cite{hafner2019dream}.
\subsubsection{Transition Network}
\begin{figure}[h]
    \centering
    \includegraphics[width=0.4\columnwidth]{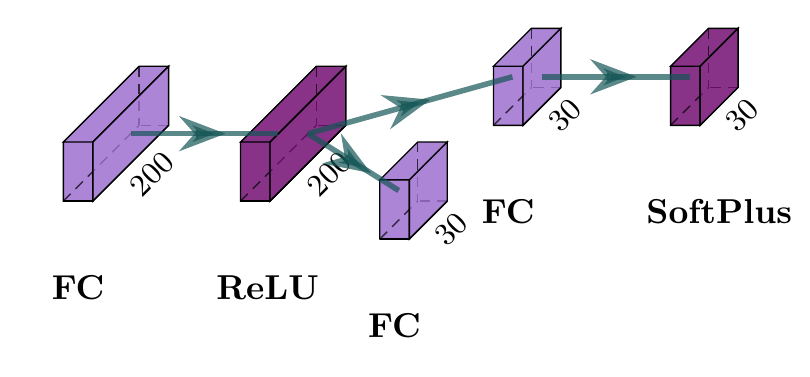}
    \caption{\small The network modeling $p_{\theta}(s_{t}^c|h_{t})$. }
    \label{fig:deter-to-stoch}
    \vspace{-0.3cm}
\end{figure}

We use a GRU module to model the deterministic transition function $g_{\theta}$. The dimension of $h_t$ is $200$ in both Multi-Modal Natural Mujoco and for Table Wiping tasks. We use a multi-layer perceptrons to model $p_{\theta}(s_{t}^c|h_{t})$. $p_{\theta}(s_{t}^c|h_{t})$ is modeled as a Gaussian with a diagonal covariance matrix. The multi-layer perceptron takes in $h_t$ as an input and outputs the mean and variance of the $p_{\theta}(s_{t}^c|h_{t})$ (Fig.\ref{fig:deter-to-stoch}).

\subsubsection{Inference Networks}
\begin{figure}[htb]
  \centering
    \includegraphics[width=0.8\columnwidth]{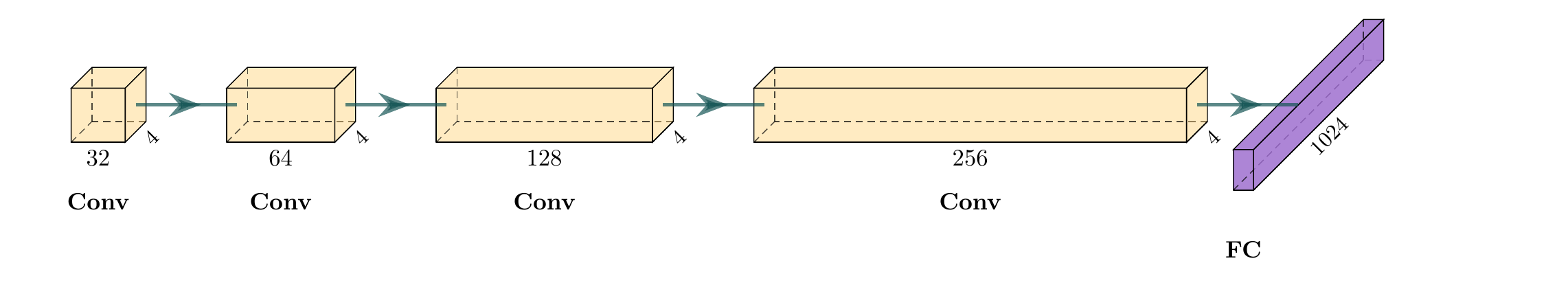}
      \caption{Embedding networks for RGB image (or depth image)}
      \label{fig:embed_image}
      \vspace{-0.3cm}
\end{figure}

\begin{figure}[htb]
  \centering
    \includegraphics[width=0.7\columnwidth]{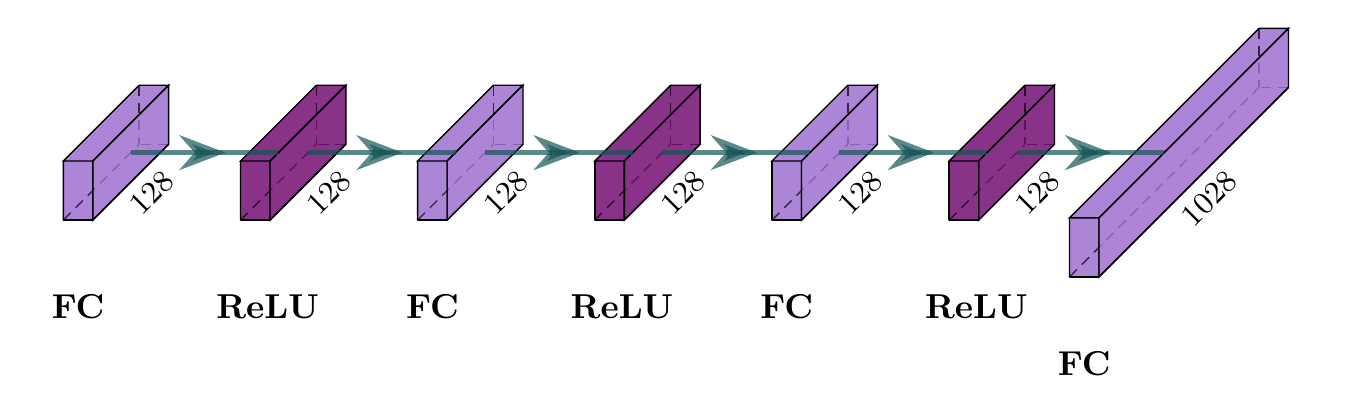}
      \caption{Embedding networks for tactile}
      \label{fig:embed_tactile}
      \vspace{-0.3cm}
\end{figure}

\begin{figure}[htb]
  \centering
      \includegraphics[width=0.7\columnwidth]{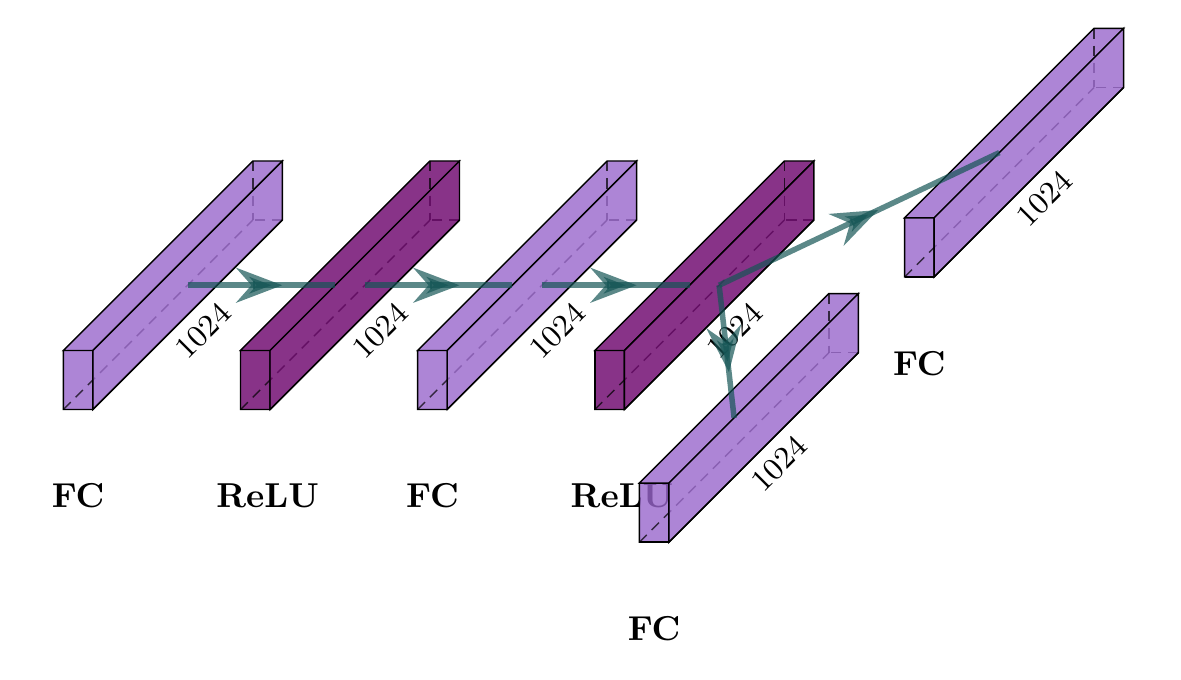}
      \caption{Network that maps extracted features to mean and variance of $q(s_t^f|x_t^m)$}
      \label{fig:embed2p}
      \vspace{-0.3cm}
\end{figure}

\begin{figure}[htb]
  \centering
      \includegraphics[width=0.4\columnwidth]{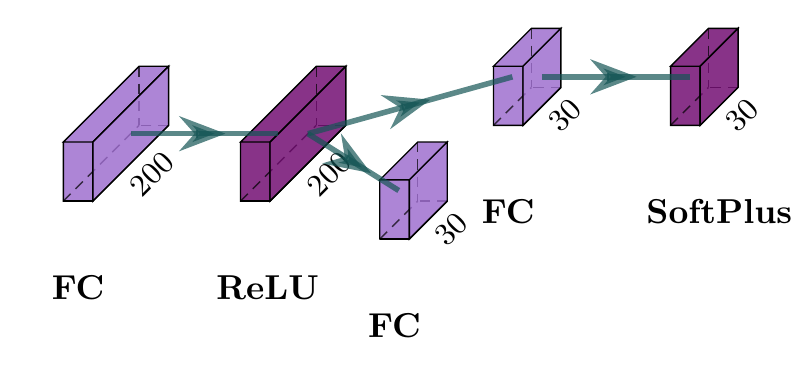}
      \caption{Network that takes in the concatenation of $s_t^f,h_t$ and outputs the mean and variance of $q(s_t^c|s_t^f,h_t)$}
      \label{fig:RSSM}
      \vspace{-0.3cm}
\end{figure}

We model $q(s_t^f|x_t^m)$ as Gaussian distributions with a diagonal covariance matrix. For each modality, we first use a network to extract features from raw data (Fig. \ref{fig:embed_image} or Fig. \ref{fig:embed_tactile}) and then use another network to map this features to the mean and variance of $q(s_t^f|x_t^m)$ (Fig.~\ref{fig:embed2p}). The dimension of $s_t^f$ is $1024$ in Multi-Modal Natural Mujoco tasks and $256$ in the Table Wiping task. $q(s_t^c|s_t^f,h_t)$ is modeled as multi-layer perceptron, which takes in the concatenation of $s_t^f, h_t$ as an input and outputs the mean and variance of $q(s_t^c|s_t^f,h_t)$ (Fig.\ref{fig:RSSM}). The dimension of $s_t^c$ is $30$ in both Multi-Modal Natural Mujoco and for Table Wiping tasks.

\subsubsection{Actor and Value Networks}
\begin{figure}[!htbp]
    \centering
    \includegraphics[width=0.95\columnwidth]{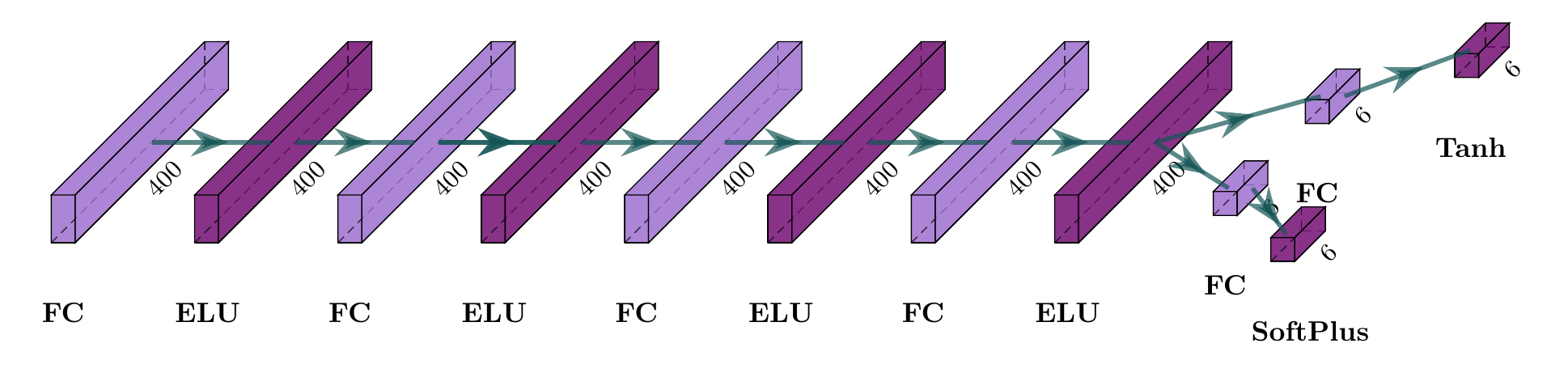}
    \caption{\small Actor network that models $\pi(a|z)$. }
    \label{fig:actor-net}
    \vspace{-0.3cm}
\end{figure}

\begin{figure}[!htbp]
    \centering
    \includegraphics[width=0.8\columnwidth]{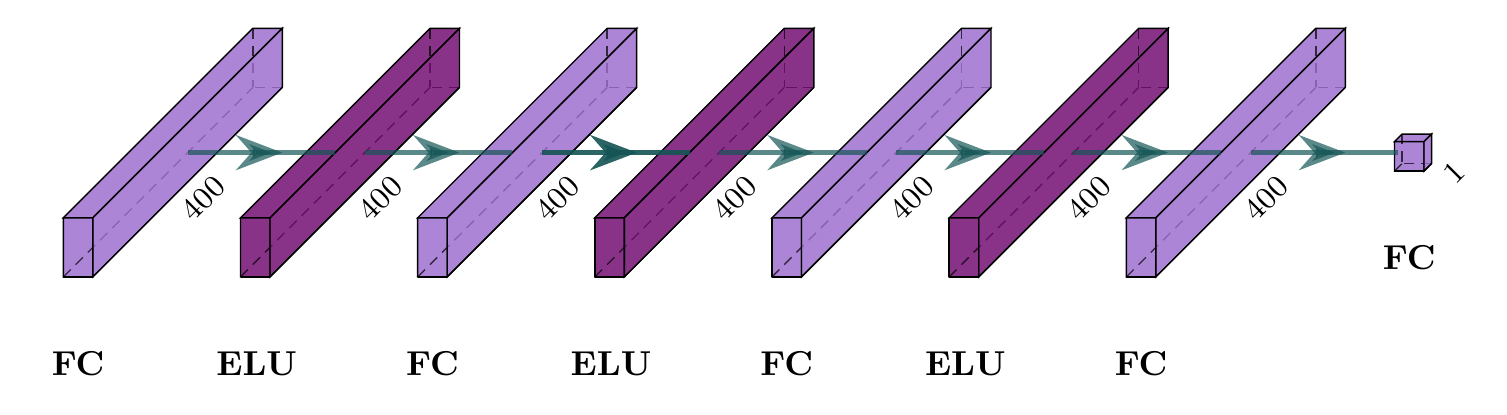}
    \caption{\small Value network that models value function $V_{\lambda}(z)$. }
    \label{fig:value-net}
    \vspace{-0.3cm}
\end{figure}

We used a multi-layer perceptron to model actor $\pi(a|z)$ and value function. We modelled the actor $\pi(a|z)$ as a Gaussian distribution with a diagonal covariance matrix. The actor networks takes latent states $z$ as input and outputs the mean and variance for $\pi(a|z)$ (Fig.\ref{fig:actor-net}). Similarly, the value network takes latent states $z$ as input and outputs the values for value function (Fig.\ref{fig:value-net}).

\subsection{Baseline Models}
\begin{figure}[!htbp]
    \centering
    \includegraphics[width=0.99\columnwidth]{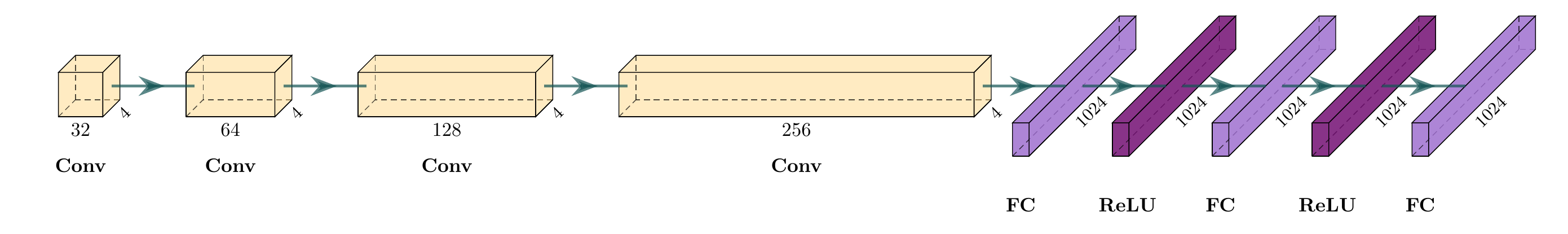}
    \caption{Networks that extract features from RGB image or depth image.}
    \label{fig:encoder_image_cat}
    \vspace{-0.3cm}
\end{figure}

\begin{figure}[!htbp]
    \centering
      \includegraphics[width=0.99\columnwidth]{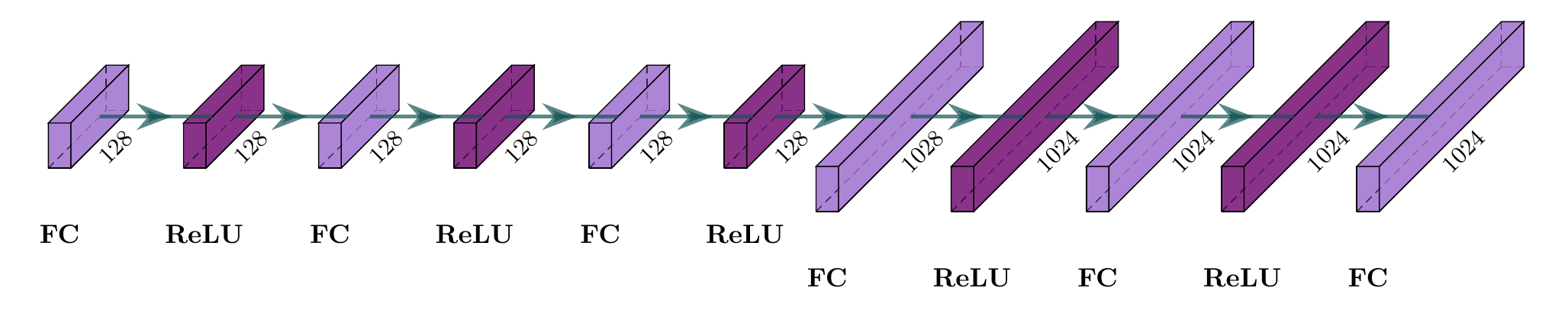}
      \caption{Networks that extract features from tactile.}
      \label{fig:encoder_tactile_cat}
      \vspace{-0.3cm}
\end{figure}

Similar network structures were used in the baseline models. However, instead the product-of-experts (PoE), we use the networks in Fig.~\ref{fig:encoder_image_cat} and Fig.~\ref{fig:encoder_tactile_cat} to first extract features from different modalities, which are contacted and fed into another network (Fig.\ref{fig:RSSM}) to obtain the mean and variance of $q(s_t^c|s_t^f,h_t)$. The dimension of $h_t$ and $s_t^f$ are the same as in MuMMI. In Multi-Modal Natural Mujoco tasks, the $s_t^f$ is of dimension $3072=1024\times3$ (the concatenation of feature vectors extracted from three modalities---RGB image, depth image and tactile). In the Table Wiping task, the $s_t^f$ is of dimension $512=256\times2$, which is the concatenation of feature vectors extracted from two modalities (RGB image from workspace camera and RGB image from robot camera).  

\begin{figure}[!htbp]
    \centering
    \includegraphics[width=0.7\columnwidth]{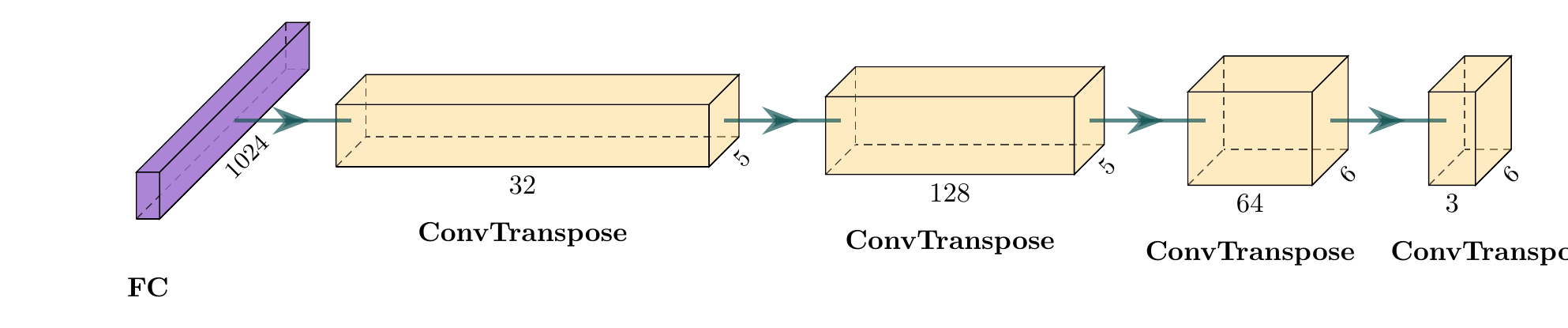}
    \caption{Decoder network for RGB image.}
    \label{fig:decoder_rgbimage}
    \vspace{-0.3cm}
\end{figure}

\begin{figure}[!htbp]
    \centering
    \includegraphics[width=0.7\columnwidth]{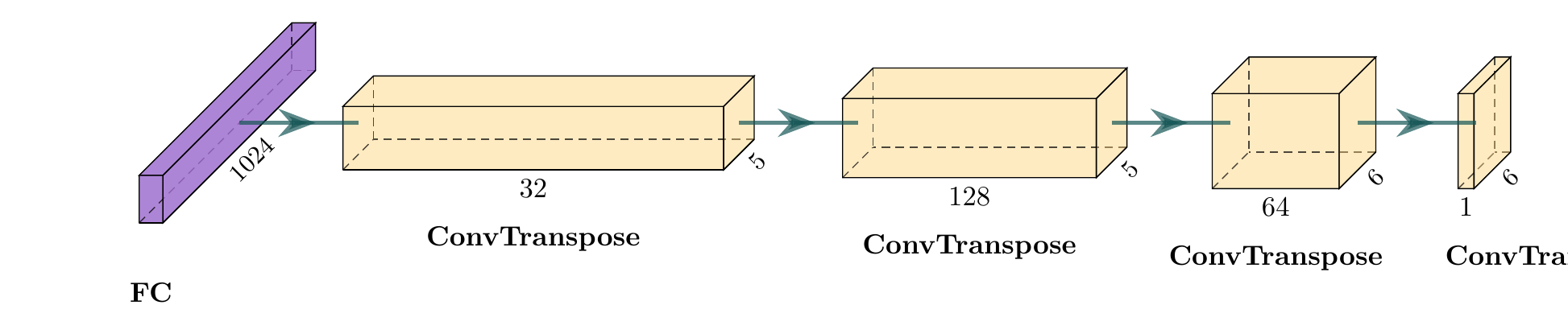}
    \caption{Decoder network for depth image.}
    \label{fig:decoder_depthimage}
    \vspace{-0.3cm}
\end{figure}

\begin{figure}[!htbp]
    \centering
      \includegraphics[width=0.7\columnwidth]{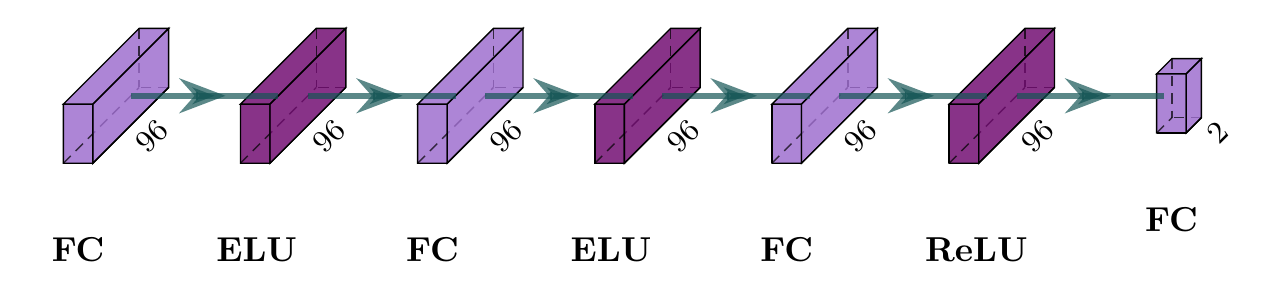}
      \caption{Decoder network for tactile with $2$ channel.}
      \label{fig:decoder_tactile}
      \vspace{-0.3cm}
\end{figure}

For Dreamer \cite{hafner2019dream}, we use decoder networks to model $p_\theta(x_t^m|z_t)$, which is assumed to be Gaussian with diagonal covariance in our experiments. The decoder networks takes in latent state as an input and outputs the mean $p_\theta(x_t^m|z_t)$ (Fig.~\ref{fig:decoder_rgbimage}, Fig.~\ref{fig:decoder_depthimage} and Fig.~\ref{fig:decoder_tactile}). The variance of $p_\theta(x_t^m|z_t)$ is set as $1.0$ for all modalities.

\subsection{Additional Results}
\begin{figure}[!htbp]
    \centering
    \includegraphics[width=0.4\columnwidth]{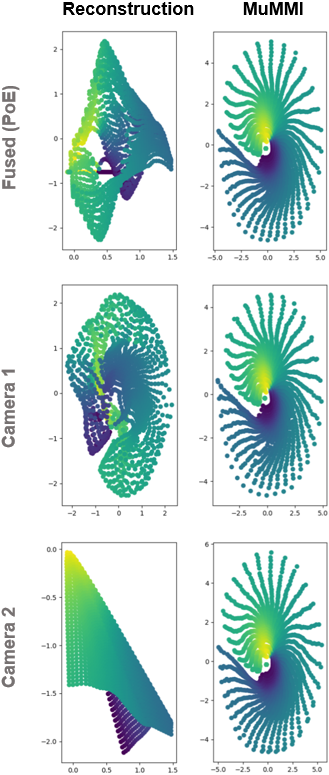}
    \caption{The robot in a 2D world with two sensory modalities: a camera mounted in the front and a camera mounted behind. The figures show the latent space learnt by reconstruction and MuMMI.}
    \label{fig:toy_exp_imgimg}
\end{figure}

\begin{figure}[!htbp]
    \centering
      \includegraphics[width=0.4\columnwidth]{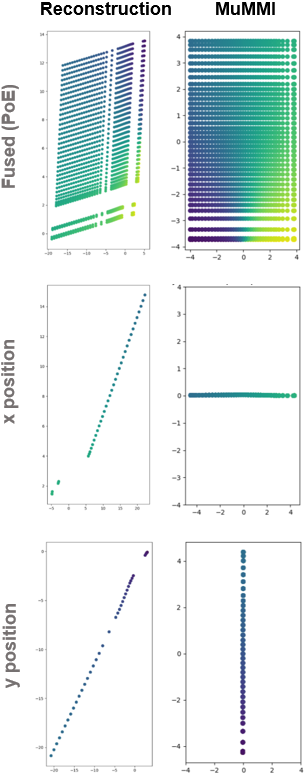}
      \caption{The robot in a 2D world with two independent sensory modalities: $x$ position of the robot and $y$ position. The figures show the latent space learnt by reconstruction and MuMMI.}
      \label{fig:toy_exp_pospos}
\end{figure}

Additional results for the toy example can be seen in Fig.~\ref{fig:toy_exp_imgimg} and Fig.~\ref{fig:toy_exp_pospos}. In Fig.~\ref{fig:toy_exp_imgimg}, we can see that using MuMMI results in a consistent representation  among two camera but using the reconstruction loss does not. Also, if the two modalities are independent (the $x$ and $y$ positions of the robot), MuMMI can still learn a structured latent space. (Fig.\ref{fig:toy_exp_pospos}).

\newpage

\end{document}